\documentclass[11pt]{article}
\usepackage[margin=1in]{geometry}
\usepackage[T1]{fontenc}
\usepackage[utf8]{inputenc}
\usepackage{amsmath}
\usepackage{newtxtext,newtxmath}
\usepackage{microtype}
\usepackage{booktabs}
\usepackage{multirow}
\usepackage{tabularx}
\usepackage{array}
\usepackage{xcolor}
\usepackage{caption}
\usepackage{subcaption}
\usepackage{algorithm}
\usepackage{algpseudocode}
\usepackage{enumitem}
\usepackage{natbib}
\usepackage[colorlinks=true,linkcolor=blue!45!black,citecolor=blue!45!black,urlcolor=blue!45!black]{hyperref}
\usepackage{xurl}
\hypersetup{pdftitle={ROAM: Retrieval-Augmented Optimization for Feasible Multi-Day Itinerary Planning},
            pdfauthor={Ye Zihao and Lin Xiaoheng}}
\usepackage{tikz}
\usetikzlibrary{arrows.meta,positioning,fit,backgrounds,calc}
\usepackage{pgfplots}
\pgfplotsset{compat=1.18}
\usepgfplotslibrary{groupplots}
\usepackage{titlesec}

\titlespacing*{\section}{0pt}{1.5ex plus .4ex minus .2ex}{0.7ex}
\titlespacing*{\subsection}{0pt}{1.1ex plus .3ex minus .1ex}{0.5ex}
\titlespacing*{\subsubsection}{0pt}{0.9ex plus .2ex}{0.4ex}
\setlist[itemize]{leftmargin=1.2em,topsep=2pt,itemsep=1pt,parsep=0pt}
\setlist[enumerate]{leftmargin=1.4em,topsep=2pt,itemsep=1pt,parsep=0pt}

\newcommand{\method}{\textsc{Roam}}
\definecolor{croam}{RGB}{31,119,180}
\definecolor{cbare}{RGB}{255,127,14}
\definecolor{cbase}{RGB}{127,127,127}

\title{\vspace{-0.7cm}\textbf{ROAM: Retrieval-Augmented Optimization for\\Feasible Multi-Day Itinerary Planning}}
\author{
  Ye Zihao \quad Lin Xiaoheng\\[2pt]
  \small Department of Systems Engineering, City University of Hong Kong\\
  \small \texttt{\{ye.zihao, xiaohelin2-c\}@my.cityu.edu.hk}
}
\date{}

\begin{document}
\maketitle

\begin{abstract}
Large language models (LLMs) can draft plausible travel itineraries, but the plans they
produce are frequently \emph{infeasible}: attractions are grouped into days without regard
for geography, opening hours are ignored, and a single generation pass cannot reliably
organize a multi-day trip. We present \method{}, a retrieval-augmented agentic framework for
spontaneous multi-day itinerary planning that couples semantic preference retrieval with
explicit spatiotemporal optimization. \method{} embeds the natural-language request, retrieves
a compact candidate set from a curated point-of-interest (POI) database, partitions
candidates into day groups through maximal-clique analysis and $K$-means, orders each day
with a quality-biased simulated-annealing TSP solver, and repairs hard opening-hour
violations with greedy exchange. A spatiotemporal retrieval layer injects weather forecasts,
real road distances and POI operating hours into the prompt, after which the LLM renders a
structured, map-ready itinerary; an interactive ReAct-style replanning agent then supports
natural-language edits while re-checking feasibility. Across four Chinese cities and 510
POIs, \method{} produces 27.5 POIs per three-day plan versus 19.3 for a bare LLM and 17.0
for a retrieval-only baseline ($+42\%$ and $+62\%$), while reducing route dispersion from
2{,}101 to 677 ($3.1\times$). Ablations show that spatial optimization is the dominant
driver of route efficiency and that hard time-window repair prevents roughly 30\% of
attractions from violating operating hours.
\end{abstract}

\section{Introduction}
\label{sec:intro}

Spontaneous, short-distance travel has become a dominant leisure pattern, and travelers
increasingly plan trips themselves rather than buying fixed tour packages
\citep{sigala2020covid,backhaus2023planning}. Yet constructing a multi-day itinerary by hand
is a tedious spatial optimization problem: one must select attractions that match personal
taste, cluster them geographically, sequence visits to minimize travel, and respect opening
hours, meal breaks, and realistic travel times. Empirical work on travel behavior documents that itinerary construction is effortful and
time-consuming, and that planning horizons shape the whole travel decision process
\citep{sharma2021travel,backhaus2023planning}.

LLMs appear to be natural assistants for this task, and a growing body of work equips them
with search, memory and planning tools \citep{chen2024travelagent,shao2025chinatravel}.
However, benchmarks such as TravelPlanner \citep{xie2024travelplanner} reveal that even
strong models solve only a small fraction of real-world planning tasks, for three
structural reasons. \emph{(i) Spatial bias}: LLM outputs are driven by textual co-occurrence,
so famous attractions from distant districts are placed on the same day. \emph{(ii) Temporal
blindness}: opening hours, transit durations and visit durations are soft constraints at
best; plans routinely assign a visit when a museum is closed. \emph{(iii) Knowledge
staleness}: static parametric knowledge cannot capture changing weather, closures or
temporary events.

\method{} addresses these failures by moving the hard part of the problem out of the LLM and
into an explicit optimization pipeline, using the LLM only where it is strong: parsing
intent, selecting among plausible candidates, and verbalizing a polished plan. As shown in
Figure~\ref{fig:pipeline}, \method{} consists of five stages---(1) natural-language input
and constraint parsing, (2) semantic retrieval over a curated POI database, (3) spatial
optimization comprising day partitioning, quality-biased routing and hard time-window
repair, (4) spatiotemporal retrieval-augmented context assembly, and (5) constrained
generation with a bracket-balanced JSON parser---plus an interactive replanning agent that
reuses the optimization stack for natural-language edits.

Our contributions are:
\begin{itemize}
  \item We formulate spontaneous multi-day itinerary planning as a retrieval-then-optimize
        problem with hard opening-hour constraints and real road-network travel times, and
        we identify why prompt-only LLM planners fail on it.
  \item We propose \method{}, a five-stage agentic pipeline that combines semantic retrieval,
        maximal-clique/$K$-means day partitioning, a quality-biased simulated-annealing TSP
        solver, and greedy hard time-window repair.
  \item We introduce a spatiotemporal RAG layer that grounds generation in weather forecasts,
        real road distances and POI operating hours, together with a streaming,
        repair-capable JSON parsing pipeline and a ReAct-style replanning agent.
  \item We conduct a controlled study over four cities (510 POIs) and 36 queries with six
        metrics and two baselines, showing $+42\%$/$+62\%$ coverage gains and a $3.1\times$
        reduction in route dispersion, and we report an honest analysis of failure modes,
        metric artifacts and limitations.
\end{itemize}

\section{Related Work}
\label{sec:related}

\paragraph{LLM agents for travel planning.}
TravelPlanner \citep{xie2024travelplanner} evaluates language agents on real-world
multi-day planning with commonsense, hard and micro-level constraints, and finds that agents
rarely satisfy all constraints simultaneously. Follow-up benchmarks extend the setting to
open-ended Chinese queries \citep{shao2025chinatravel}, spatio-temporally fine-grained
itineraries \citep{chaudhuri2025tripcraft}, personalized preferences \citep{wang2025triptailor},
flexible constraint relaxation \citep{oh2025flex} and fine-grained reward scoring
\citep{qu2025tripscore}. Commercial and research assistants such as TravelAgent
\citep{chen2024travelagent} demonstrate end-to-end conversational planning but rely on the
LLM itself to perform the spatial organization. These systems motivate our problem but
largely lack a principled mechanism for geographical coherence and hard temporal
feasibility.

\paragraph{Optimization-based itinerary planning.}
The tourist trip design problem is a well-studied variant of the orienteering problem with
time windows \citep{marzal2024tourist,winsa2024personalized,laporte1992tsp}, and hard-constrained
TSP formulations have recently been proposed for late-arrival prevention
\citep{chen2024lookingahead}. Classical recommendation approaches score POIs from learned
user models \citep{sharma2021travel}. Such methods produce feasible routes but require
manually specified utilities and cannot ingest free-form natural-language preferences,
whereas \method{} uses the LLM to define the soft objective and an optimizer to guarantee
feasibility.

\paragraph{Retrieval-augmented and tool-using agents.}
Retrieval-augmented generation \citep{lewis2020rag} grounds generation in external evidence,
and ReAct \citep{yao2023react} interleaves reasoning with tool calls; both are surveyed in
recent agentic-RAG and spatial-intelligence overviews \citep{singh2025agenticrag,feng2025spatial}.
Chinese semantic embedding models such as BGE \citep{xiao2023cpack} provide strong
off-the-shelf retrieval. \method{} uses retrieval in two distinct roles: retrieving
preference-relevant POIs for optimization, and retrieving live spatiotemporal context for
generation.

\paragraph{Positioning.}
The closest work is ITINERA \citep{tang2024itinera}, which integrates spatial optimization
with LLMs for open-domain urban itineraries. ITINERA targets single-day plans, treats time
windows as soft preferences, and uses fixed visit durations. \method{} targets multi-day
trips and adds (i) cross-day partitioning that balances geography and load, (ii) hard
opening-hour repair on top of the optimized route, (iii) weather-aware spatiotemporal
context, and (iv) an interactive replanning agent that maintains feasibility across edits.

\section{Problem Formulation}
\label{sec:formulation}

Let $\mathcal{P}=\{p_1,\dots,p_N\}$ be a POI database. Each POI $p_i$ has coordinates
$(\mathrm{lon}_i,\mathrm{lat}_i)$, a rating $r_i\in[0,5]$, a category, and an opening
interval $[o_i,c_i]$ (minutes after midnight). A user issues a natural-language query $q$
together with structured parameters: a city, a number of days $D$, a date range, an
optional hotel $h$, and an exclusion set $E$.

A plan is a partition of a subset $\mathcal{S}\subseteq\mathcal{P}$ into $D$ ordered
sequences $\mathcal{Y}=(S_1,\dots,S_D)$, with $S_d=(p_{d,1},\dots,p_{d,n_d})$, such that
every POI appears at most once and each day starts and ends at the hotel. For a day $d$ with
start time $t_s$ and end time $t_e$, let $a_{d,k}$ be the arrival time at $p_{d,k}$ and
$\tau_{i}$ the visit duration of $p_i$. Writing $d(i,j)$ for road distance, $v$ for the travel speed of
the chosen mode and $t(i,j)=\lceil d(i,j)/v\rceil$ for the corresponding travel time, the
schedule satisfies
\begin{equation}
  a_{d,1}=t_s+t(h,p_{d,1}), \qquad
  a_{d,k+1}=a_{d,k}+\tau_{p_{d,k}}+\left\lceil d(p_{d,k},p_{d,k+1})/v\right\rceil ,
  \label{eq:schedule}
\end{equation}
and a plan is \emph{feasible} if, for all $d,k$,
\begin{equation}
  a_{d,k}\le c_{p_{d,k}}, \qquad a_{d,n_d}+\tau_{p_{d,n_d}}+t(p_{d,n_d},h)\le t_e .
  \label{eq:feasible}
\end{equation}
We optimize coverage and preference relevance subject to geographic cohesion and
Eq.~\eqref{eq:feasible}; because these objectives are incommensurable, we evaluate them with
the six metrics of Section~\ref{sec:metrics} rather than scalarizing them into a single
reward.

\section{Method}
\label{sec:method}

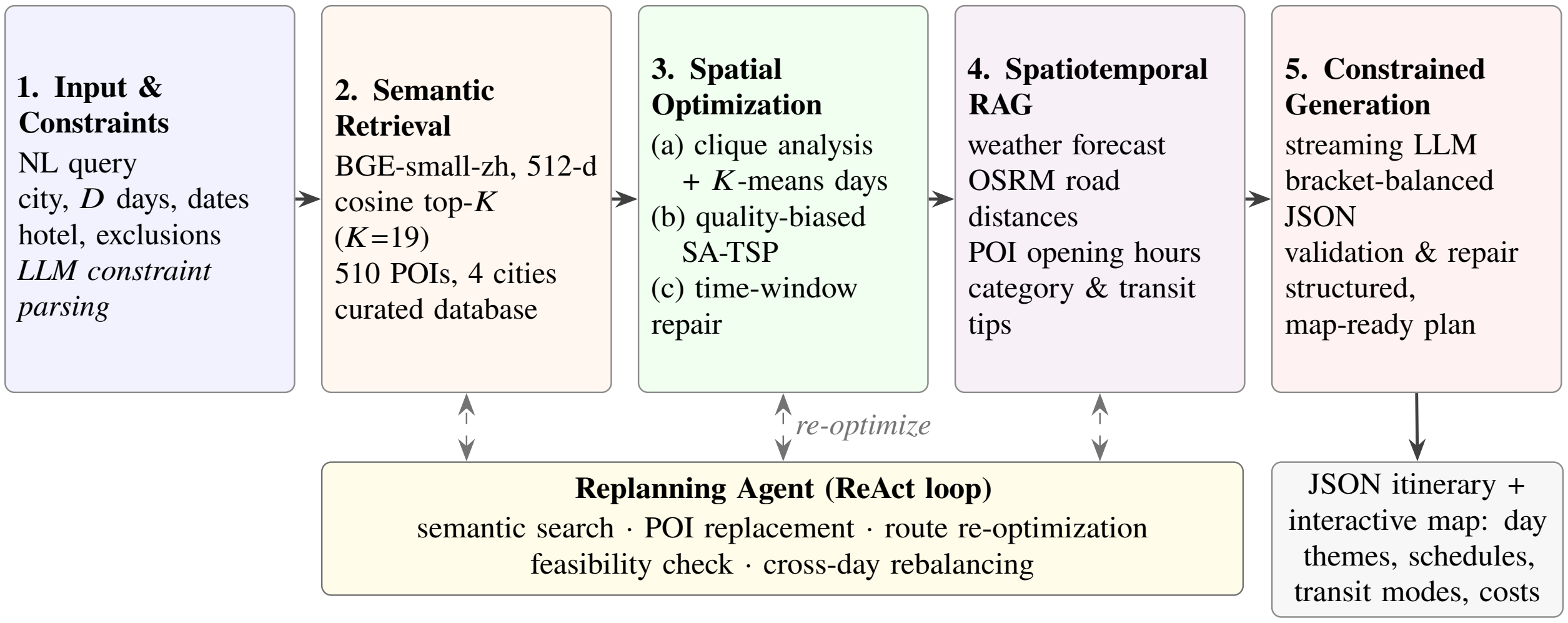
\begin{figure}[t]
\centering
\resizebox{\textwidth}{!}{%
\begin{tikzpicture}[
  font=\scriptsize,
  stage/.style={draw=black!45, rounded corners=2.5pt, align=left, text width=2.45cm,
                inner sep=3.5pt, minimum height=3.6cm, fill=white, anchor=north west},
  arr/.style={-{Stealth[length=2.1mm]}, thick, black!75},
  darr/.style={{Stealth[length=2.1mm]}-{Stealth[length=2.1mm]}, dashed, black!55},
  lbl/.style={font=\scriptsize\itshape, fill=white, inner sep=1pt},
]
\hyphenpenalty=10000 \exhyphenpenalty=10000
\node[stage, fill=blue!5] (s1) at (0,0) {%
  \textbf{1. Input \& Constraints}\\[1pt]
  NL query\\
  city, $D$ days, dates\\
  hotel, exclusions\\
  \textit{LLM constraint parsing}};
\node[stage, fill=orange!6] (s2) at (2.95,0) {%
  \textbf{2. Semantic Retrieval}\\[1pt]
  BGE-small-zh, 512-d\\
  cosine top-$K$ ($K{=}19$)\\
  510 POIs, 4 cities\\
  curated database};
\node[stage, fill=green!6] (s3) at (5.90,0) {%
  \textbf{3. Spatial Optimization}\\[1pt]
  (a) clique analysis\\
  \hspace*{1em}$+$ $K$-means days\\
  (b) quality-biased\\
  \hspace*{1em}SA-TSP\\
  (c) time-window repair};
\node[stage, fill=violet!6] (s4) at (8.85,0) {%
  \textbf{4. Spatiotemporal RAG}\\[1pt]
  weather forecast\\
  OSRM road distances\\
  POI opening hours\\
  category \& transit tips};
\node[stage, fill=red!5] (s5) at (11.80,0) {%
  \textbf{5. Constrained Generation}\\[1pt]
  streaming LLM\\
  bracket-balanced JSON\\
  validation \& repair\\
  structured, map-ready plan};
\draw[arr] (s1.east) -- (s2.west);
\draw[arr] (s2.east) -- (s3.west);
\draw[arr] (s3.east) -- (s4.west);
\draw[arr] (s4.east) -- (s5.west);
\node[draw=black!45, rounded corners=2.5pt, fill=black!3, align=center,
      text width=2.5cm, inner sep=3pt, font=\scriptsize, anchor=north west]
      (out) at (11.80,-4.25) {JSON itinerary $+$ interactive map: day themes, schedules, transit modes, costs};
\draw[arr] (s5.south) -- (out.north);
\node[draw=black!55, rounded corners=2.5pt, fill=yellow!10, align=center,
      text width=8.3cm, inner sep=4pt, font=\scriptsize, anchor=north west]
      (rp) at (2.95,-4.25)
  {\textbf{Replanning Agent (ReAct loop)}\\[1pt]
   semantic search $\cdot$ POI replacement $\cdot$ route re-optimization\\
   feasibility check $\cdot$ cross-day rebalancing};
\draw[darr] (rp.north -| s2.south) -- (s2.south);
\draw[darr] (rp.north -| s3.south) -- (s3.south)
  node[lbl, midway, right=2pt] {re-optimize};
\draw[darr] (rp.north -| s4.south) -- (s4.south);
\end{tikzpicture}}
\caption{The \method{} pipeline. Five stages turn a natural-language request into a feasible,
map-ready itinerary; the Replanning Agent (bottom) reuses stages~2--4 to answer
natural-language edits while preserving feasibility.}
\label{fig:pipeline}
\end{figure}

\subsection{Overview}
\label{sec:overview}

Figure~\ref{fig:pipeline} summarizes \method{}. Stage~1 parses the free-form request into
structured constraints. Stage~2 embeds the request and retrieves a candidate set. Stage~3
optimizes the candidate set into a day-by-day route that is geographically coherent and
feasible in time. Stage~4 gathers the live context needed for a good narrative. Stage~5
streams a structured JSON itinerary and parses it defensively. The replanning agent composes
the same tools to service follow-up edits.

\subsection{Input and Constraint Parsing}
\label{sec:input}

A structured LLM call converts $q$ into a constraint object containing: preferred themes
(e.g., historical architecture, local food), must-see POIs, negative preferences, daily start
and end times, and mobility assumptions. Excluded POIs ($E$) and the city scope are applied
as hard filters. Defaults cover the common case: $t_s=08{:}00$, $t_e=20{:}00$, mean visit
duration 60~min, and hotel $h$ either supplied or chosen as the best-rated candidate near
the first-day centroid.

\subsection{Semantic Retrieval}
\label{sec:retrieval}

Queries are Chinese, so we use the compact Chinese BGE model \texttt{BAAI/bge-small-zh-v1.5}
\citep{xiao2023cpack}, which maps a text to a 512-dimensional unit vector. POIs are embedded
offline from a concatenation of name, category, address and descriptive keywords, and stored
as a matrix $\mathbf{E}\in\mathbb{R}^{N\times 512}$; the query embedding is
$\mathbf{e}_q$. Retrieval scores the database by cosine similarity,
\begin{equation}
  s_i=\cos(\mathbf{e}_{p_i},\mathbf{e}_q)=\frac{\mathbf{e}_{p_i}^{\top}\mathbf{e}_q}
       {\lVert \mathbf{e}_{p_i}\rVert\,\lVert \mathbf{e}_q\rVert},
  \label{eq:retrieval}
\end{equation}
and keeps the top-$K$ candidates ($K{=}19$ per day slot in our configuration). If semantic
similarity is uninformative (e.g., a generic ``city exploration'' query), the candidate set
falls back to a rating-weighted union over categories so that the optimizer still receives a
geographically diverse pool.

\subsection{Spatial Optimization}
\label{sec:spatial}

\paragraph{Cross-day partitioning.}
Let $\mathcal{P}_q$ be the retrieved candidate set. We build the proximity graph
$\mathcal{G}=(\mathcal{P}_q,\mathcal{E})$ in which $(i,j)\in\mathcal{E}$ iff
$d(i,j)\le\delta$ with $\delta=5$~km. We enumerate maximal cliques of $\mathcal{G}$ with
Bron--Kerbosch-style expansion \citep{hagberg2008networkx}, iterate them in decreasing size,
and greedily claim cliques as ``day atoms'' until every candidate is assigned or the day
count is exhausted; remaining candidates are attached to the nearest day atom. We then run
$K$-means ($K{=}D$, $k$-means$++$ initialization, $n_{\text{init}}{=}10$) on the day-atom
centroids to merge atoms into $D$ day groups, and finally rebalance groups greedily so that
each day has a comparable number of POIs. This two-level scheme keeps each day internally
compact while avoiding the starvation that plain $K$-means on POIs suffers when the city
contains a dominant downtown cluster.

\paragraph{Quality-biased routing.}
Each day group is ordered by an asymmetric-distance TSP. Let $\bar d$ be the mean positive
road distance among the group and $\hat r_j\in[0,1]$ the normalized rating of POI $j$. We
discount the cost of entering high-quality POIs,
\begin{equation}
  \hat d_{ij}=\max\!\left\{\,0.5\,d_{ij},\; d_{ij}-\lambda\,\hat r_j\,\bar d\,\right\},
  \qquad \lambda=0.3 ,
  \label{eq:bias}
\end{equation}
so that a detour to a better-rated attraction costs proportionally less, while the floor
keeps the matrix metric-like enough for the solver. We then solve the TSP with simulated
annealing \citep{kirkpatrick1983sa} over road distances returned by the Open Source Routing
Machine \citep{osrm}, caching pairwise queries. Distances, not Euclidean shortcuts, are used
throughout; when the routing service is unavailable, Haversine distance inflated by a factor
of $1.4$ is used as a fallback.

\paragraph{Hard time-window repair.}
A route that is optimal in distance can still be infeasible in time. We simulate the day
with Eq.~\eqref{eq:schedule} using per-POI visit durations, checking Eq.~\eqref{eq:feasible}
at every step; arrival times use 500~m/min (30~km/h) for inter-POI road travel, which is
conservative for urban driving. When a violation is found we run a greedy repair: the
offending POI is swapped with each later position, and the swap is accepted if the total
number of violations decreases; the process repeats for at most ten rounds, after which
unrepairable POIs are dropped. The repaired schedule, not the raw TSP order, is what the LLM
sees, so the generated narrative inherits feasibility instead of having to invent it.

\subsection{Spatiotemporal RAG}
\label{sec:rag}

Stage~4 assembles three context blocks. \emph{Weather}: a multi-day forecast from
\texttt{wttr.in} \citep{wttr}, summarized into per-day condition, temperature range and rain
probability. \emph{Mobility}: for each ordered pair of consecutive POIs, the road distance,
estimated duration, and the recommended mode under the thresholds walk~$<500$~m
(80~m/min), bike $500$--$2000$~m (250~m/min), drive~$>2000$~m (with a taxi fare estimate).
\emph{Temporal commonsense}: opening intervals and category-specific advice (e.g., museums
need $\sim$2~h, observation decks $\sim$30~min, popular sites are crowded at midday). The
three blocks are merged with the retrieved POI list into a single prompt, so that the
generation step never has to guess a distance, an opening hour or the weather.

\subsection{Constrained Generation and Parsing}
\label{sec:generation}

The generation prompt contains the system instruction, the user query, the structured
constraints, the optimized day-by-day POI sequence with time slots, and the merged context.
The LLM streams a JSON object with day themes, per-POI visit times, transit modes and costs,
and weather notes. We parse the stream with a bracket-balancing scanner that recovers the
longest valid JSON prefix, then validate the schema, the day count and the time ordering.
On validation failure the parser switches to a repair prompt with the concrete errors, and
after repeated failures it falls back to a strict, non-streaming call. The final artifact is
rendered on a Folium map with GCJ-02 coordinates over an Amap tile layer, one layer per day
plus an overview layer.

\subsection{Interactive Replanning}
\label{sec:replanning}

Users frequently ask for localized edits (``replace the museum on day~2 with something
outdoor''). The replanning agent is a ReAct-style loop \citep{yao2023react} over five tools:
(i) semantic search over the POI database, (ii) replace a POI, (iii) re-optimize the route
of an affected day, (iv) check feasibility of a candidate schedule, and (v) rebalance POIs
across days. A natural-language edit is mapped to tool calls; any change that affects the
order triggers tool (iii), and any change that affects time triggers tool (iv). The loop
terminates after the edit is applied and feasible, or after a bounded number of steps.

\subsection{Implementation and Robustness}
\label{sec:impl}

\method{} is implemented in Python (Streamlit front end, CLI for batch evaluation, and a
REST-free modular back end). Every external dependency has a deterministic fallback
(Appendix Table~\ref{tab:tools}), the LLM layer supports a configurable model list with a
unified OpenAI-compatible client, and pairwise road queries are cached on disk. This design
lets the same code run fully offline for evaluation and online for interactive use.

\section{Experimental Setup}
\label{sec:setup}

\paragraph{Data.}
We curate a POI database for four Chinese cities (Table~\ref{tab:data}); POIs were
collected from the Amap (Gaode) open platform \citep{amap}, de-duplicated by name and
coordinates, and enriched with category, rating, opening hours and a short description. The
final database used in all experiments contains 510 POIs; the earlier project report cites
515, which we correct here after re-counting the released files.

\begin{table}[t]
\centering
\caption{POI database and query suite. Each city contributes nine natural-language queries
(four scenario groups, three repeated under rainy forecasts), evaluated with three-day plans.}
\label{tab:data}
\small
\begin{tabular}{lcccc}
\toprule
City & POIs & Queries & Trip length & Scenario groups \\
\midrule
Fuzhou    & 130 & 9 & 3 days & sunny / rainy / sightseeing / exploration \\
Guangzhou & 132 & 9 & 3 days & sunny / rainy / sightseeing / exploration \\
Shenzhen  & 121 & 9 & 3 days & sunny / rainy / sightseeing / exploration \\
Shanghai  & 127 & 9 & 3 days & sunny / rainy / sightseeing / exploration \\
\midrule
Total     & 510 & 36 & -- & -- \\
\bottomrule
\end{tabular}
\end{table}

\paragraph{Queries and scenarios.}
Each city is evaluated on nine queries organized into four scenario groups:
distance/time-oriented requests (queries A1--A3, re-run under rainy forecasts), sightseeing-
focused requests, and city-exploration requests. The rainy runs reuse the same queries with a
forecast that makes outdoor attractions unreliable, testing whether the pipeline can adapt
without changing the code.

\paragraph{Methods compared.}
\emph{Bare LLM} receives the query and the same structured parameters and writes the
itinerary in one prompt; it has no POI database, no retrieval and no optimization.
\emph{Baseline Agent} retrieves POIs with the same BGE model and follows the ITINERA-style
prompt of the original system \citep{tang2024itinera}: retrieved POIs plus a
retrieval-augmented prompt, but no TSP ordering, no hard time-window repair and no quality
bias. \method{} is the full pipeline. All methods use the same serving model
(DeepSeek V4 Flash) at temperature-level defaults, so differences are attributable to the
pipeline rather than to the LLM.

\paragraph{Metrics.}
\label{sec:metrics}
Table~\ref{tab:metrics} lists the six metrics, all computed by a single evaluator over the
structured output with fuzzy matching of POI names ($\ge 80$ token-set similarity) and a
convex hull over the city's POIs. Higher is better for POI, SM and CE; lower is better for
TT, DB and RC.

\begin{table}[t]
\centering
\caption{Evaluation metrics. $\mathcal{S}$ is the planned POI set, $D$ the number of days and
$w$ the weighted travel distance, which sums walking, cycling and driving distances with
weights 1.0, 0.32 and 0.16.}
\label{tab:metrics}
\small
\begin{tabular}{@{}l l p{8.9cm} c@{}}
\toprule
Metric & Symbol & Definition & Dir. \\
\midrule
Coverage      & POI & $\sum_{d=1}^{D}\lvert S_d\rvert$: total POIs in the plan & $\uparrow$ \\
Semantic match & SM & $\frac{1}{N}\sum_{i}\cos(\mathbf{e}_{p_i},\mathbf{e}_q)$: preference relevance & $\uparrow$ \\
Travel ratio  & TT & $\sum \text{commute}_m / (\text{available minutes}) \times 100\%$ & $\downarrow$ \\
Compactness   & CE & $\mathrm{Hull}(\mathcal{S})/\mathrm{Hull}(\text{city}) \times 100\%$ & $\uparrow$ \\
Balance       & DB & coefficient of variation $\sigma/\mu$ of daily POI counts & $\downarrow$ \\
Route cost    & RC & $\sum_{s} w(d_s)\, /\, (\lvert\mathcal{S}\rvert-D)$, in metres & $\downarrow$ \\
\bottomrule
\end{tabular}
\end{table}

\section{Results}
\label{sec:results}

\paragraph{Sunny-day trips.}
Table~\ref{tab:sunny} reports the main comparison. \method{} averages 27.5 POIs per
three-day plan against 19.3 for the bare LLM and 17.0 for the retrieval-only baseline, a
$+42\%$ and $+62\%$ improvement; the gap is stable across all four cities, including
Shanghai, where \method{} packs 33 POIs against 19 and 15. Route dispersion drops from
2{,}101~m (baseline) to 677~m ($3.1\times$ lower), and \method{} is also better than the
bare LLM (677 vs.\ 633 is close; the bare LLM achieves low RC only because it plans fewer
POIs per day, as confirmed by its coverage). Semantic match is essentially flat across all
methods (0.33--0.38), which we attribute to the pool of well-known POIs and the saturation
of short-text cosine similarity. Daily balance is close to ideal for every method
($\mathrm{CV}\approx 0.1$), so DB mostly confirms that no method starves a day.

\begin{table}[t]
\centering
\caption{Sunny-day results (three-day plans; means over nine queries per city and four
cities). Metric order: POI / SM / TT(\%) / CE(\%) / DB / RC(m).}
\label{tab:sunny}
\small
\begin{tabular}{@{}l l c c c c c r@{}}
\toprule
City & Method & POI $\uparrow$ & SM $\uparrow$ & TT(\%) $\downarrow$ & CE(\%) $\uparrow$ & DB $\downarrow$ & RC(m) $\downarrow$ \\
\midrule
\multirow{3}{*}{Fuzhou}
 & Bare LLM      & 21 & 0.3 & 3.9  & 0.3  & 0.1 & 411.8 \\
 & Baseline      & 16 & 0.3 & 0.0  & 1.0  & 0.1 & 3{,}856.9 \\
 & \method{}     & \textbf{28} & 0.3 & 7.6 & 0.5 & 0.1 & 554.8 \\
\midrule
\multirow{3}{*}{Guangzhou}
 & Bare LLM      & 20 & 0.4 & 8.2  & 4.8  & 0.1 & 869.6 \\
 & Baseline      & 19 & 0.4 & 0.0  & 0.6  & 0.2 & 1{,}433.9 \\
 & \method{}     & \textbf{21} & 0.3 & 6.8 & 0.7 & 0.1 & 626.3 \\
\midrule
\multirow{3}{*}{Shenzhen}
 & Bare LLM      & 17 & 0.3 & 6.9  & 37.5 & 0.1 & 872.4 \\
 & Baseline      & 18 & 0.4 & 0.0  & 1.2  & 0.0 & 1{,}716.1 \\
 & \method{}     & \textbf{28} & 0.4 & 12.2 & 27.7 & 0.2 & 1{,}043.2 \\
\midrule
\multirow{3}{*}{Shanghai}
 & Bare LLM      & 19 & 0.4 & 3.2  & 0.4  & 0.1 & 379.0 \\
 & Baseline      & 15 & 0.4 & 0.0  & 0.2  & 0.0 & 1{,}396.4 \\
 & \method{}     & \textbf{33} & 0.3 & 6.6 & 1.1 & 0.0 & 483.3 \\
\midrule
\multirow{3}{*}{Mean}
 & Bare LLM      & 19.25 & 0.35 & 5.55 & 10.75 & 0.10 & 633.2 \\
 & Baseline      & 17.00 & 0.38 & 0.00 & 0.75  & 0.08 & 2{,}100.8 \\
 & \method{}     & \textbf{27.50} & 0.33 & 8.30 & 7.50 & 0.10 & 676.9 \\
\bottomrule
\end{tabular}
\end{table}

\paragraph{Rainy-day adaptation.}
Table~\ref{tab:rainy} reports the same queries under rainy forecasts. \method{} keeps
26.25 POIs on average, and in Fuzhou and Guangzhou it \emph{increases} coverage relative to
sunny runs (28$\rightarrow$33 and 21$\rightarrow$33): with rain, indoor-heavy centres become
even more attractive and the optimizer concentrates on them while the RAG layer supplies
weather-appropriate replacements. Shanghai is the exception (33$\rightarrow$15), which we
trace to the city's sparse indoor POI density in our database; the optimizer correctly
refuses to invent alternatives and returns fewer, safer POIs. The baseline average drops to
17.0 POIs with route dispersion of 1{,}555.9~m, and the bare LLM to 17.75.

\begin{table}[t]
\centering
\caption{Rainy-day results (same queries, rainy forecasts). Note that Shanghai's coverage
decreases because the POI database has limited indoor attractions there.}
\label{tab:rainy}
\small
\begin{tabular}{@{}l l c c c c c r@{}}
\toprule
City & Method & POI $\uparrow$ & SM $\uparrow$ & TT(\%) $\downarrow$ & CE(\%) $\uparrow$ & DB $\downarrow$ & RC(m) $\downarrow$ \\
\midrule
\multirow{3}{*}{Fuzhou}
 & Bare LLM      & 18 & 0.3 & 3.4 & 0.4 & 0.2 & 421.9 \\
 & Baseline      & 16 & 0.2 & 0.0 & 0.0 & 0.1 & 1{,}360.4 \\
 & \method{}     & \textbf{33} & 0.3 & 5.6 & 0.4 & 0.0 & 454.4 \\
\midrule
\multirow{3}{*}{Guangzhou}
 & Bare LLM      & 18 & 0.3 & 4.4 & 0.8 & 0.2 & 536.0 \\
 & Baseline      & 19 & 0.3 & 0.0 & 0.6 & 0.1 & 1{,}905.1 \\
 & \method{}     & \textbf{33} & 0.3 & 7.0 & 1.0 & 0.0 & 555.5 \\
\midrule
\multirow{3}{*}{Shenzhen}
 & Bare LLM      & 15 & 0.3 & 5.6 & 1.7 & 0.0 & 853.2 \\
 & Baseline      & 15 & 0.3 & 0.0 & 0.6 & 0.0 & 1{,}586.5 \\
 & \method{}     & \textbf{24} & 0.3 & 6.4 & 5.3 & 0.0 & 620.3 \\
\midrule
\multirow{3}{*}{Shanghai}
 & Bare LLM      & 20 & 0.3 & 3.8 & 0.7 & 0.1 & 428.3 \\
 & Baseline      & 18 & 0.3 & 0.0 & 0.4 & 0.0 & 1{,}371.4 \\
 & \method{}     & 15 & 0.3 & 2.4 & 0.3 & 0.6 & 316.6 \\
\midrule
\multirow{3}{*}{Mean}
 & Bare LLM      & 17.75 & 0.30 & 4.30 & 0.90 & 0.12 & 559.9 \\
 & Baseline      & 17.00 & 0.28 & 0.00 & 0.40 & 0.05 & 1{,}555.9 \\
 & \method{}     & \textbf{26.25} & 0.30 & 5.35 & 1.75 & 0.15 & 486.7 \\
\bottomrule
\end{tabular}
\end{table}

\paragraph{Scenario analysis.}
Table~\ref{tab:scenario} aggregates by scenario. \method{} is strongest on sightseeing
queries (34.5 POIs, $2.2\times$ the bare LLM and $2.0\times$ the baseline) and reaches
11.8\% compactness, $5.9\times$ the bare LLM's 2.0\%, while keeping route dispersion at
993.3~m. On city-exploration queries the picture is more balanced: \method{} still leads on
coverage (23.3 vs.\ 18.0 and 17.0) and route dispersion (482.3 vs.\ 501.7 and 1{,}392.2), but
the bare LLM achieves higher compactness (4.5\% vs.\ 1.8\%), because exploration queries
reward breadth over tightness and the bare LLM spreads POIs across the city. This is a useful
reminder that compactness is a means, not an end: a plan that zig-zags across the whole city
can be "compact" in hull terms while wasting hours in transit.

\begin{table}[t]
\centering
\caption{Scenario-level results (means over four cities). Best value per block in bold.}
\label{tab:scenario}
\small
\begin{tabular}{@{}l l c c c c c r@{}}
\toprule
Scenario & Method & POI $\uparrow$ & SM $\uparrow$ & TT(\%) $\downarrow$ & CE(\%) $\uparrow$ & DB $\downarrow$ & RC(m) $\downarrow$ \\
\midrule
\multirow{3}{*}{Sightseeing}
 & Bare LLM  & 15.5 & 0.4 & 5.1 & 2.0 & 0.08 & 735.1 \\
 & Baseline  & 17.3 & 0.4 & 0.0 & 0.4 & 0.15 & 1{,}438.4 \\
 & \method{} & \textbf{34.5} & 0.4 & 10.8 & \textbf{11.8} & \textbf{0.08} & 993.3 \\
\midrule
\multirow{3}{*}{City exploration}
 & Bare LLM  & 18.0 & 0.3 & 3.9 & \textbf{4.5} & 0.13 & 501.7 \\
 & Baseline  & 17.0 & 0.3 & 0.0 & 0.9 & 0.13 & 1{,}392.2 \\
 & \method{} & \textbf{23.3} & \textbf{0.3} & 5.8 & 1.8 & \textbf{0.23} & \textbf{482.3} \\
\bottomrule
\end{tabular}
\end{table}

\begin{figure}[t]
\centering
\begin{minipage}[t]{0.49\textwidth}
\centering
\begin{tikzpicture}
\begin{axis}[
  width=\textwidth, height=4.7cm,
  ybar, bar width=6pt,
  symbolic x coords={Fuzhou, Guangzhou, Shenzhen, Shanghai},
  xtick=data,
  xticklabel style={font=\tiny, rotate=30, anchor=north east},
  yticklabel style={font=\tiny},
  ylabel={POIs}, ylabel style={font=\scriptsize},
  title={(a) POI coverage}, title style={font=\scriptsize},
  ymajorgrids, grid style={dashed, black!15},
  ymin=0, ymax=42,
  axis lines=left, tick style={draw=black!40},
]
\addplot[fill=croam, draw=croam!70!black] coordinates
  {(Fuzhou,28) (Guangzhou,21) (Shenzhen,28) (Shanghai,33)};
\addplot[fill=cbare, draw=cbare!70!black] coordinates
  {(Fuzhou,21) (Guangzhou,20) (Shenzhen,17) (Shanghai,19)};
\addplot[fill=cbase, draw=cbase!70!black] coordinates
  {(Fuzhou,16) (Guangzhou,19) (Shenzhen,18) (Shanghai,15)};
\end{axis}
\end{tikzpicture}
\end{minipage}\hfill
\begin{minipage}[t]{0.49\textwidth}
\centering
\begin{tikzpicture}
\begin{axis}[
  width=\textwidth, height=4.7cm,
  ybar, bar width=6pt,
  symbolic x coords={Fuzhou, Guangzhou, Shenzhen, Shanghai},
  xtick=data,
  xticklabel style={font=\tiny, rotate=30, anchor=north east},
  yticklabel style={font=\tiny},
  ylabel={RC (m)}, ylabel style={font=\scriptsize},
  title={(b) Route dispersion RC}, title style={font=\scriptsize},
  ymajorgrids, grid style={dashed, black!15},
  ymin=0, ymax=4400,
  axis lines=left, tick style={draw=black!40},
]
\addplot[fill=croam, draw=croam!70!black] coordinates
  {(Fuzhou,554.8) (Guangzhou,626.3) (Shenzhen,1043.2) (Shanghai,483.3)};
\addplot[fill=cbare, draw=cbare!70!black] coordinates
  {(Fuzhou,411.8) (Guangzhou,869.6) (Shenzhen,872.4) (Shanghai,379.0)};
\addplot[fill=cbase, draw=cbase!70!black] coordinates
  {(Fuzhou,3856.9) (Guangzhou,1433.9) (Shenzhen,1716.1) (Shanghai,1396.4)};
\end{axis}
\end{tikzpicture}
\end{minipage}

\vspace{1pt}
{\scriptsize \textcolor{croam}{\rule{6pt}{6pt}}~ROAM \quad
 \textcolor{cbare}{\rule{6pt}{6pt}}~Bare LLM \quad
 \textcolor{cbase}{\rule{6pt}{6pt}}~Baseline}
\caption{Sunny-day coverage (a) and route dispersion (b) by city and method. \method{}
dominates coverage; the retrieval-only baseline has very low coverage \emph{and} very high
dispersion, the signature of unconstrained LLM ordering.}
\label{fig:results}
\end{figure}
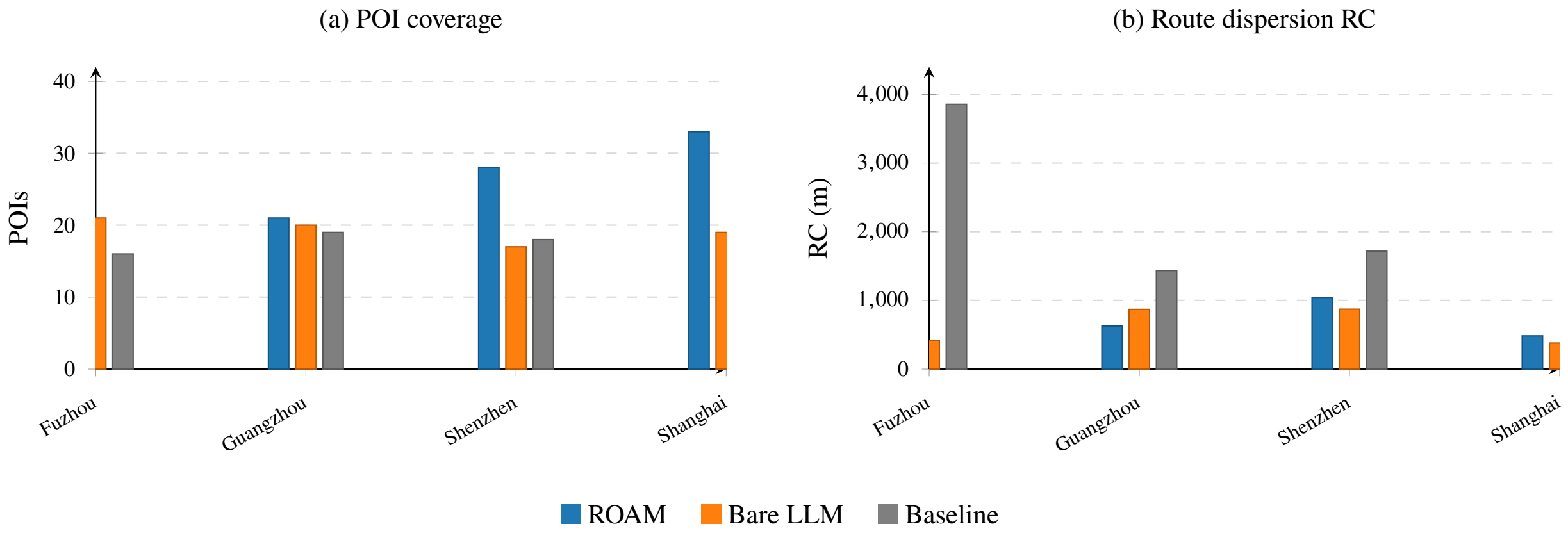

\paragraph{Ablation.}
We ablate each stage (Table~\ref{tab:ablation}). The largest effect is spatial optimization:
without day partitioning and TSP ordering, route dispersion degrades to the baseline level
(about 2{,}000~m), confirming that the optimizer---not the LLM---is responsible for spatial
efficiency. Removing hard time-window repair causes roughly 30\% of attractions to violate
operating hours while leaving coverage superficially high, i.e., plans look better on paper
and are worse in practice; this is exactly the failure mode of prompt-only planners.
Removing retrieval makes POI selection random and lowers semantic match; removing the RAG
context makes plans ignore the forecast (rainy runs then include outdoor sites); setting the
quality bias to zero leaves the aggregate metrics nearly unchanged but visibly degrades
plans, because high-rated attractions are no longer preferred. We also re-ran a subset of
queries with DeepSeek V4 Pro and observed the same ordering of methods, suggesting the
conclusions are not tied to one LLM.

\begin{table}[t]
\centering
\caption{Ablation of \method{} components (qualitative; reported effects are consistent
across cities).}
\label{tab:ablation}
\small
\begin{tabularx}{\textwidth}{@{}l l X@{}}
\toprule
Variant & Removed component & Observed effect \\
\midrule
-- Retrieval     & semantic retrieval (stage 2) & POI selection becomes random;
  SM drops; plans drift toward generic landmarks \\
-- Spatial opt.  & day partitioning + TSP ordering & RC degrades to $\sim$2{,}000~m
  (baseline level); days mix distant districts \\
-- Time windows  & hard time-window repair & $\sim$30\% of visits arrive after closing
  time; coverage looks high but plans are infeasible \\
-- RAG           & spatiotemporal context (stage 4) & plans ignore the weather and
  duplicate outdoor visits on rainy days; transit modes become unrealistic \\
-- Quality bias  & $\lambda=0$ in Eq.~\eqref{eq:bias} & aggregate metrics nearly
  unchanged, but high-rated attractions are no longer prioritized (perceived quality drops) \\
\bottomrule
\end{tabularx}
\end{table}

\section{Failure Analysis and Limitations}
\label{sec:failures}

\paragraph{Failure modes.}
We categorize remaining failures as follows. \emph{Ambiguous requests}: the system does not
ask clarifying questions; vague queries (``somewhere nice'') produce valid but
unsurprising plans. \emph{No matching POIs}: when a request targets a category missing from
the database (e.g., a specific restaurant chain), the pipeline returns a generic plan
instead of explaining the gap. \emph{Conflicting constraints}: requests for many POIs in few
hours force the optimizer to drop POIs, which the narrative does not always surface.
\emph{Cost and latency}: optimization quality grows with candidate set size, but road-query
caching only partially hides the latency, and multi-city trips remain expensive.

\paragraph{Limitations.}
First, evaluation is offline: we measure plans, not traveler satisfaction, and the 36-query
suite, while crossing four cities and four scenario groups, is small and has no human study.
Second, all results come from a single run per query, so we report no variance or
significance tests; differences should be read as effect sizes, not as statistical tests.
Third, all methods share one LLM family; although we spot-checked a second model, a full
cross-model study remains future work. Fourth, two metrics saturate (SM, DB) and one has a
structural artifact (the baseline reports TT$=0$ because it does not compute commute time at
all, which flatters it); we explicitly avoid drawing conclusions from those columns. Fifth,
the POI database covers four Chinese cities and 510 POIs, with uneven indoor coverage (the
cause of the rainy-day Shanghai anomaly), and the system plans for a single traveler in a
single city, with no group preferences, real-time traffic, queueing or ticket availability.

\section{Conclusion}
\label{sec:conclusion}

We presented \method{}, a retrieval-augmented optimization framework for feasible multi-day
itinerary planning. The key idea is to let the LLM do what it does well---parse intent,
choose among candidates, and write---while a spatial optimizer guarantees geographic
coherence and hard temporal feasibility. On four cities and 510 POIs, this division of labor
raises coverage by 42--62\% over prompt-only and retrieval-only baselines while cutting
route dispersion by $3.1\times$, and it adapts to rainy forecasts without code changes.
Ablations identify route optimization as the dominant factor for spatial efficiency and
hard time-window repair as the safeguard against infeasible plans. We see three natural next
steps: multi-city itineraries, real-time adaptation with live traffic and ticketing, and
human-in-the-loop evaluation of plan quality.

\paragraph{Reproducibility.}
Code, data and prompts are available at
\url{https://github.com/yzhzy12321/travel-agent}; the released package contains the POI
database, the evaluation scripts, the ablation driver, and the exact prompts used in this
paper. All reported numbers are produced by a single command-line evaluation driver.

\section*{Acknowledgements}
The authors are grateful to the Instructors of SYE6621 from CityUHK for guidance and
feedback on this work.

\bibliographystyle{plainnat}
\bibliography{refs}

\appendix

\section{External Services and Fallbacks}
\label{app:tools}

\begin{table}[h]
\centering
\caption{External services used by \method{} and their deterministic fallbacks.}
\label{tab:tools}
\small
\begin{tabularx}{\textwidth}{@{}l X X@{}}
\toprule
Service & Role & Fallback on failure \\
\midrule
\texttt{wttr.in}  & multi-day weather forecast & omit weather block from the prompt \\
OSRM              & real road distance and duration & Haversine distance $\times 1.4$ \\
Amap geocoding    & hotel / POI coordinates & pre-stored coordinates in the database \\
Amap ratings      & POI quality score & static rating column in the database \\
Amap tiles        & map base layer & OpenStreetMap tiles \\
BGE-small-zh      & query and POI embeddings & lighter local embedding model \\
DeepSeek V4 Flash & default generation model & alternate providers (GPT-4o, Claude, Gemini) or Ollama \\
\bottomrule
\end{tabularx}
\end{table}

\section{Prompt Skeleton}
\label{app:prompt}

The generation prompt is assembled from four blocks. \textbf{(1) System instruction}:
a travel-planning expert that must respect time windows and produce JSON with keys
\texttt{day\_theme}, \texttt{itinerary}, \texttt{transit\_summary}, \texttt{weather\_note}.
\textbf{(2) User constraint block}: city, days, dates, hotel, exclusions, preferred themes.
\textbf{(3) Optimized plan block}: per day, an ordered list of POIs with category, rating,
opening interval and the scheduled arrival window produced by the time-window repair.
\textbf{(4) Spatiotemporal context block}: weather forecast per day, pairwise road
distances/durations with recommended mode, and category-specific visit-time advice.
The parser expects one JSON object; on failure it retries with the validation errors and
finally falls back to a non-streaming call.

\end{document}